\documentclass[letterpaper, 10 pt, conference]{ieeeconf}  % Comment this line out if you need a4paper

\IEEEoverridecommandlockouts                              % This command is only needed if 
\usepackage{graphicx} % for \resizebox and graphics
\usepackage{amsmath} % assumes amsmath package installed
\usepackage{amssymb}  % assumes amsmath package installed
\usepackage{booktabs}
\usepackage{dblfloatfix}
\usepackage{multirow}
\usepackage{tabularx}
\usepackage{xcolor}

\usepackage[font=footnotesize]{caption}
\usepackage{subcaption}
\usepackage{mathtools} % for \mathclap

\title{\LARGE \bf
Partial Motion Imitation for Learning Cart Pushing \\ with Legged Manipulators
}

\author{Mili Das$^{1*}$, Morgan Byrd$^{1}$, Donghoon Baek$^{1}$, and Sehoon Ha$^{1}$
\thanks{$^{1}$Georgia Institute of Technology, Atlanta, GA, 30308, USA}%
\thanks{*Correspondence to mdas45@gatech.edu}
}

\begin{document}

\maketitle{}
\thispagestyle{empty}
\pagestyle{empty}

%%%%%%%%%%%%%%%%%%%%%%%%%%%%%%%%%%%%%%%%%%%%%%%%%%%%%%%%%%%%%%%%%%%%%%%%%%%%%%%%
\begin{abstract}
Loco-manipulation is a key capability for legged robots to perform practical mobile manipulation tasks, such as transporting and pushing objects, in real-world environments. However, learning robust loco-manipulation skills remains challenging due to the difficulty of maintaining stable locomotion while simultaneously performing precise manipulation behaviors. This work proposes a partial imitation learning approach that transfers the locomotion style learned from a locomotion task to cart loco-manipulation. A robust locomotion policy is first trained with extensive domain and terrain randomization, and a loco-manipulation policy is then learned by imitating only lower-body motions using a partial adversarial motion prior. We conduct experiments demonstrating that the learned policy successfully pushes a cart along diverse trajectories in IsaacLab and transfers effectively to MuJoCo. We also compare our method to several baselines and show that the proposed approach achieves more stable and accurate loco-manipulation behaviors. 
% \sehoon{mention no vision in abstract, intro, and problem formulation}

% This electronic document is a ÒliveÓ template. The various components of your paper [title, text, heads, etc.] are already defined on the style sheet, as illustrated by the portions given in this document.

\end{abstract}

\section{INTRODUCTION}

% \sehoon{Mili, add citation everywhere}

% Motivation
Legged robots have recently demonstrated increasingly capable locomotion and manipulation skills in real-world environments \cite{liu2024visualwholebodycontrollegged, fu2022deepwholebodycontrollearning, portela2024learningforcecontrollegged}. In particular, legged manipulators—quadrupedal robots with manipulators mounted on top—offer a promising solution for mobile manipulation tasks in unstructured factory or home environments, such as transporting objects and assisting with everyday activities. Among these tasks, pushing everyday objects, like carts, is especially relevant for practical automation scenarios, including warehouse logistics and retail environments. Successful deployment in these settings requires coordinated loco-manipulation behaviors that maintain stable locomotion while simultaneously generating accurate manipulation forces over long time horizons.

% Challenge
However, learning robust loco-manipulation policies remains challenging. Recent approaches often rely on automated frameworks, such as deep reinforcement learning, to discover coordinated whole-body behaviors. These methods typically require extensive reward engineering and meticulous training design. The simpler task of learning stable locomotion alone often involves a large number of reward terms, sometimes exceeding 10 to 20 components, alongside extensive domain randomization to achieve robust performance across diverse environments. Extending these approaches to loco-manipulation further increases the complexity of policy learning and naively combining locomotion and manipulation objectives often leads to unstable or inefficient solutions.

% Insight
To address these complexities, we draw inspiration from self-imitation. Recent motion imitation approaches have shown promising results in humanoid~\cite{fu2024humanplushumanoidshadowingimitation} and quadrupedal control problems~\cite{li2022fastmimicmodelbasedmotionimitation}. However, such methods typically rely on large-scale reference datasets, which are not available for loco-manipulation tasks. Existing datasets, such as OMOMO~\cite{li2023object}, focus primarily on full-body human motions and are not directly applicable to legged loco-manipulation. Instead, we propose to imitate the locomotion style that emerges from a simpler task and transfer it to the more complex setting of loco-manipulation. Specifically, we preserve the lower-body motion patterns that produce stable locomotion while allowing the arm to adapt freely to the manipulation task.

\begin{figure}[t]
\centering
\includegraphics[width=0.9\linewidth]{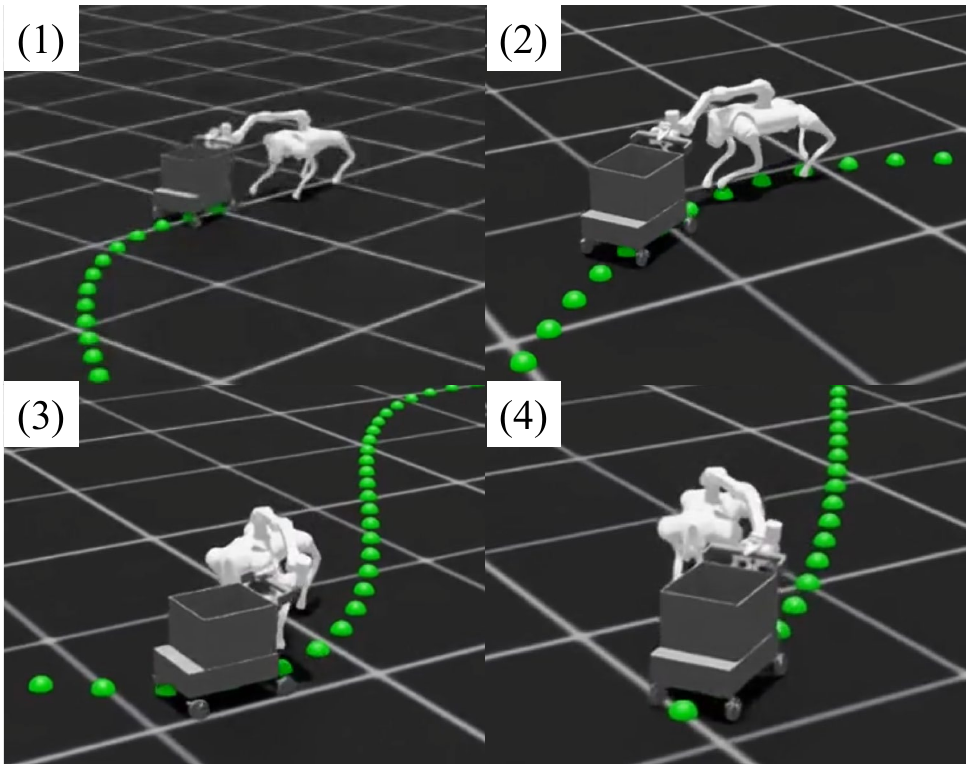}
\caption{Cart loco-manipulation along a predefined path (green dotted curve). The robot initiates contact at (1), performs straight pushing and coordinated turning maneuvers (2–4), continuously adjusting whole-body coordination to maintain stable contact and balance while tracking the target trajectory.}
\label{fig:teaser}
\vspace{-1.5em}
\end{figure}

% Method
This paper presents a novel framework for learning a robust loco-manipulation policy using partial imitation learning. We first learn a locomotion policy on rough terrain. At this stage, the locomotion policy is trained with extensive domain and terrain randomization to obtain robust and transferable walking behaviors. We then train a loco-manipulation policy by imitating only the lower-body motions using an adversarial motion prior (AMP)~\cite{Peng_2021}, which preserves the locomotion style while allowing the arm to learn effective manipulation behaviors. 
Our key insight is that full-body imitation would unnecessarily constrain manipulation behaviors, yet learning without imitation often results in unstable locomotion because the policy must simultaneously discover locomotion and manipulation strategies. By imitating only the lower-body motions, the learned policy maintains stable locomotion while enabling effective manipulation and can be combined with a standard path-following controller to push a cart along desired trajectories.
% Note that terrain randomization is essential to learn robust locomotion styles, but it is not feasible to simulate terrain variation while pushing a cart. \todo{It is not exactly clear here why the terrain randomization is needed. Either here or somewhere else in the paper it should be explicitly stated that the manipulator induces a COM shift in the robot base, which encourages a bounding gait as opposed to a trotting or walking gait (as exhibited by the No AMP ablation). Additionally, I also anticipate a reviewer may wonder why we didn't use an explicit contact timing schedule method to encourage the appropriate gait like some related work. I need some guidance on how to exactly word this/address this.}

% Results
We evaluate the proposed learning framework on a cart loco-manipulation task and observe consistently stable and reliable performance, while assuming reliable cart position estimation. When combined with a widely adopted high-level path-following controller, the learned policy successfully tracks a variety of trajectories in the IsaacLab simulation environment over long time horizons. The policy also transfers effectively across simulators, demonstrating successful sim-to-sim transfer from IsaacLab to MuJoCo. Compared to baseline approaches without imitation or with full-body imitation, our method achieves significantly improved performance and stability. 

The main contributions of this work are:
\begin{itemize}
\item We propose a novel partial imitation learning approach for robust legged loco-manipulation.

\item We demonstrate stable cart loco-manipulation through extensive simulation experiments, including strong performance in the training environment (IsaacLab) and successful sim-to-sim transfer (MuJoCo).

\item We show that the proposed approach outperforms baseline methods, including policies trained without imitation, with full-body imitation, and hierarchical RL.
% \item We propose a novel partial imitation learning approach for robust legged loco-manipulation.

% \item We demonstrate stable cart loco-manipulation through extensive simulation experiments, including baseline comparisons and sim-to-sim transfer.

% \item We validate the approach on hardware and demonstrate successful sim-to-real transfer. \morgan{We definitely do not do this.}
\end{itemize}

%%%%%%%%%%%%%%%%%%%%%%%%%%%%%%%%%%%%%%%%%%%%%%%%%%%%%%%%%%%%%%%%%%%%%%%%%%%%%%%%
\section{RELATED WORKS}

\subsection{Quadrupedal Locomotion}
% \sehoon{Review a few model-based and learning-based quadrupedal papers. You can take a look at one of our lab's paper.} 

% Model-based approaches to quadrupedal locomotion use model-predictive control formulations that solve simplified rigid-body dynamics online at high rates \cite{kim2019highlydynamicquadrupedlocomotion, gaitandtrajopti, kim2023armp}. Youm et al. \cite{Youm_2023} bridge model-based and learning-based paradigms by imitating and fine-tuning an MPC reference to obtain robust and symmetric quadrupedal gaits. Although these methods offer interpretable behavior and stability guarantees, they depend on accurate dynamics models and predefined contact schedules, limiting robustness on unstructured terrain with unknown surface properties. 

Model-based approaches are one of the key methodologies for solving quadrupedal locomotion. To do this, they typically use model-predictive control formulations that solve simplified rigid-body dynamics online at high rates~\cite{kim2019highlydynamicquadrupedlocomotion, gaitandtrajopti, kim2023armp, Youm_2023}. These methods are appealing as they offer interpretable behavior and stability guarantees. However, they depend on accurate dynamics models and predefined contact schedules, limiting robustness on unstructured terrain with unknown surface properties.
% Will add more citations

% Learning-based methods address these limitations by training policies in simulation with extensive domain randomization and transferring them to hardware \cite{tan2018simtoreallearningagilelocomotion, ha2020learningwalkrealworld, Lee_2020, rho2025unsupervised}. A key challenge is that terrain and dynamics properties are rarely observable at deployment, making the problem partially observable. Dominant strategies include teacher-student distillation of privileged terrain information \cite{Lee_2020, Miki_2022} and online adaptation modules that infer latent environment parameters from the proprioceptive history \cite{kumar2021rmarapidmotoradaptation}. Massively parallel GPU simulation has further accelerated progress by reducing training time from hours to minutes \cite{rudin2022learningwalkminutesusing}. Our work follows this proprioception-only, domain-randomized paradigm to train a reference locomotion policy, whose walking style is distilled into a loco-manipulation policy via partial adversarial imitation.

Learning-based methods address these limitations by training robust policies in simulation with extensive domain randomization and transferring them to hardware \cite{tan2018simtoreallearningagilelocomotion, ha2020learningwalkrealworld, Lee_2020, rho2025unsupervised, li2023learningagilityadaptivelegged}. Along with improving robustness via domain randomization, these methods also attempt to handle the partial observability of key state information, e.g. dynamics and terrain properties, at test time. Dominant strategies include teacher-student distillation of privileged terrain information \cite{Lee_2020, Miki_2022} and online adaptation modules that infer latent environment parameters from the proprioceptive history \cite{kumar2021rmarapidmotoradaptation, byrd-2025, Yu-RSS-17}. Our work follows the proprioception-only, domain-randomized paradigm to train a reference locomotion policy, whose walking style is distilled into a loco-manipulation policy via partial adversarial imitation.

\vspace{-0.3em}
\subsection{Legged Loco-Manipulation} 

Early work on legged loco-manipulation methods relied on model-based whole-body control over known dynamics \cite{sentiswbc, alma, sleiman2021unifiedmpcframeworkwholebody}, while recent learning-based approaches have demonstrated whole-body coordination across humanoid and quadrupedal platforms through deep controllers \cite{fu2022deepwholebodycontrollearning}, force tracking \cite{portela2024learningforcecontrollegged}, trajectory optimization \cite{liu2025opt2skillimitatingdynamicallyfeasiblewholebody}, teleoperation imitation \cite{ha2024umilegsmakingmanipulation}, and residual control \cite{byrd2026adaptmaniplearningadaptivewholebody}. Manipulating large objects with unactuated degrees of freedom and partially observed dynamics, such as carts with passive wheels, remains particularly challenging due to intermittent contacts and unmodeled frictional effects. Existing approaches address this with various methods, including explicit online planning \cite{ravan2024combiningplanningdiffusionmobility}, human demonstration priors \cite{li2025robotmoverlearninglargeobjects}, or combining model-based control with RL \cite{cheng2025rambo}. In contrast, our method learns a single end-to-end policy without planning, demonstrations, or a model, instead regularizing locomotion style using a partial adversarial motion prior derived from a reference policy.

\vspace{-0.3em}
\subsection{Adversarial Imitation Learning} 

% Adversarial imitation learning methods reduce the need for hand-crafted rewards by learning a discriminator-based training signal from expert demonstrations, in contrast to behavior cloning, which learns a policy by directly matching expert actions. Generative Adversarial Imitation Learning (GAIL) \cite{ho2016generativeadversarialimitationlearning} uses a discriminator trained to distinguish expert state-action pairs from policy-generated pairs, encouraging the policy to match expert behavior.

% High quality, deployable controllers for loco-manipulation require a large amount of expert reward tuning or expert reference data. Adversarial imitation learning methods leverage expert demonstrations to reduce the need for hand-crafted rewards by learning a discriminator-based training signal to match the policy and expert state-action distributions. Generative Adversarial Imitation Learning (GAIL) \cite{ho2016generativeadversarialimitationlearning} uses a discriminator trained to distinguish expert state-action pairs from policy-generated pairs, encouraging the policy to match expert behavior.

High quality, deployable controllers for loco-manipulation require a large amount of expert reward tuning or expert reference data. Adversarial imitation learning methods such as GAIL~\cite{ho2016generativeadversarialimitationlearning} leverage expert demonstrations to reduce the need for hand-crafted rewards by learning a discriminator-based training signal to match the policy and expert state-action distributions.

Building on this discriminator-based formulation, Adversarial Motion Priors (AMP) \cite{Peng_2021} enables adversarial imitation from state-only information (e.g., state transitions) rather than requiring expert actions and has been widely adopted for learning skills in character control \cite{merel2017learninghumanbehaviorsmotion, mu2025smpreusablescorematchingmotion}, locomotion \cite{escontrela2022adversarialmotionpriorsmake, vollenweider2022advancedskillsmultipleadversarial, wu2023learningrobustandagileleggedloco}, and whole-body control \cite{chen2026learninghumanlikebadmintonskills, wang2025physhsirealworldgeneralizablenatural} domains.

Unlike these existing works, our methodology incorporates \textit{partial AMP}, where only a selected subset of the reference agent’s joint state is imitated, and the discriminator and policy operate in the corresponding subset of the target agent’s state rather than the complete state. The most relevant comparison to our work is WASABI \cite{li2023learning}, where they leverage approximate demonstrations and reduced state information for learning agile behaviors. Our method, on the other hand, is more aligned with hierarchical control, where we separate the learning of a high quality locomotion policy from the manipulation policy learning by using partial state information.
%  \morgan{I feel like WASABI is the clearest similar work to ours. Unsure if this is the right way to show differences.}
% \sehoon{We need 25~30 papers in this section}

\section{Learning Robust Loco-manipulation using Partial Adversarial Motion Prior}

% This section presents a novel framework for learning a robust loco-manipulation policy using partial adversarial learning. We selected a legged manipulator, where we mount a WidowX AI arm \cite{widowx} on top of a Unitree Go2 quadrupedal robot \cite{unitreego2}, and aim to push a kid-sized shopping cart \cite{melissadoug_shoppingcart}.  

We propose a framework for learning a robust loco-manipulation policy using partial adversarial learning. Our platform is a legged manipulator consisting of a WidowX AI arm \cite{widowx} mounted on a Unitree Go2 quadruped \cite{unitreego2}, tasked with pushing a kid-sized shopping cart \cite{melissadoug_shoppingcart}.

% We structure the training procedure into two stages. First, we train a robust, cyclic, and symmetric locomotion reference motion policy for the legged manipulator, with maximal domain and terrain randomization. Then, we train the loco-manipulation policy while self-imitating only the lower-body reference motions using partial adversarial motion imitation. In this stage, terrain randomization is removed because it is not feasible when pushing a cart. The overview of our method is provided in Fig. \ref{fig:architecture}.

\begin{figure*}[t]
\vspace{0.4em}
\centering
\includegraphics[width=14cm]{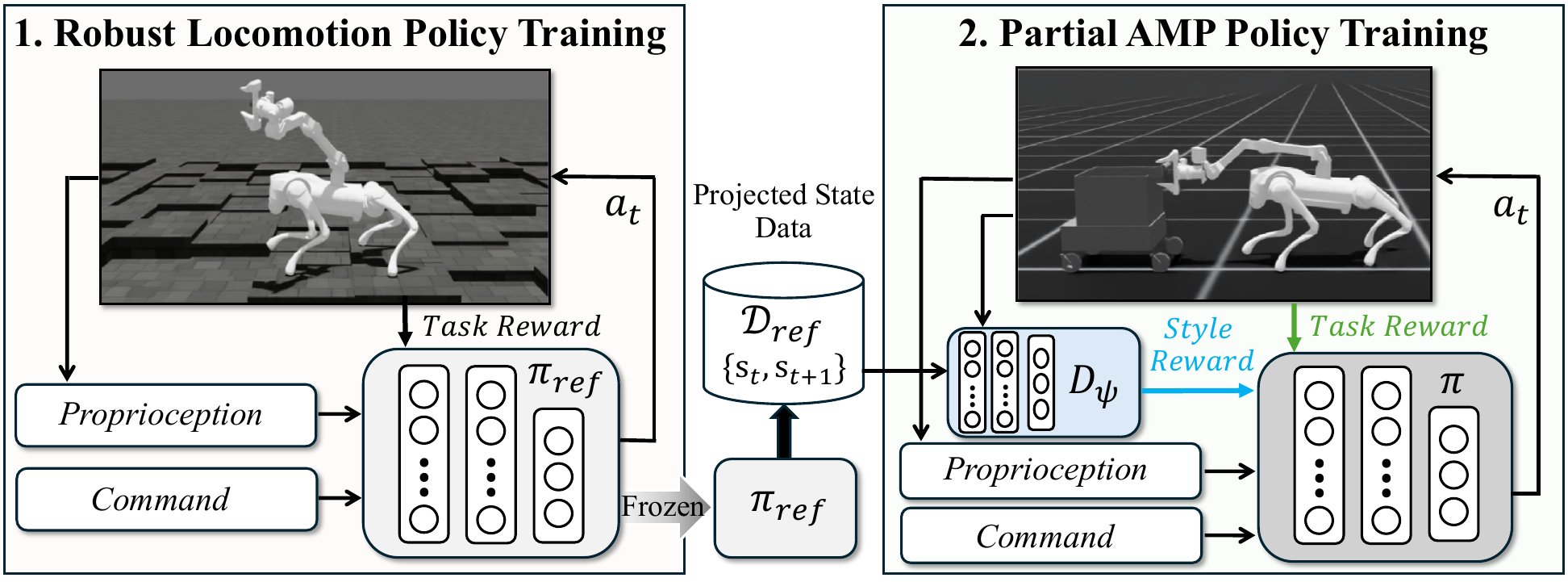}
\caption{\textbf{Architecture of the proposed framework.}
Training is performed in two stages.
In stage 1, a robust locomotion reference policy $\pi_{\text{ref}}$ is trained on rough terrain.
The resulting policy generates projected state transitions
$\mathcal{D}_{\text{ref}}=\{s_t, s_{t+1}\}$ that encode stable locomotion behaviors.
In stage 2, we train the loco-manipulation policy using partial adversarial motion priors,
where $\mathcal{D}_{\text{ref}}$ is used as the reference data for the discriminator to provide a style reward.
By imitating only the lower-body motions, the policy maintains stable locomotion while allowing flexible manipulation for cart pushing.} 
\label{fig:architecture}
\vspace{-1em}
\end{figure*}

The training procedure is divided into two stages. First, we learn a robust, cyclic, and symmetric locomotion reference policy with extensive domain and terrain randomization. Second, we train the loco-manipulation policy while self-imitating only the lower-body reference motions via partial adversarial motion imitation. Terrain randomization is removed in this stage, as uneven terrain is incompatible with stable cart pushing. An overview of the framework is shown in Fig.~\ref{fig:architecture}.

\subsection{Background: Markov Decision Process}

% We formulate the problem of learning loco-manipulation as a Markov Decision Process (MDP) \cite{sutton_reinforcement_2018}: \((S, A, f, r, \gamma)\), where \(S\) is the state space, \(A\) is the action space, \(f\) is the transition function, \(r\) is the reward function, and \(\gamma\) is the discount factor. The goal of reinforcement learning is to find parameters \(\theta\) of a policy \(\pi_{\theta}(a | s)\) that maximize the expected discounted return,  
% \begin{equation}
% J(\theta) = \mathbb{E}_{\tau \sim \pi_{\theta}} [\Sigma^{T - 1}_{t=0} \gamma^t r(s_t,a_t)],
% \end{equation}
% where \(\tau = (s_0, a_0, s_1, ...)\) denotes a trajectory induced by \(\pi_\theta\).

% When the agent cannot observe the full state, the problem becomes a Partially Observable MDP (POMDP) \cite{kaelbling1998planning}, defined by \((S,A, O, f, \Omega, r, \gamma)\), where \(O\) is the observation space and \(\Omega: S \rightarrow O\) is the observation function. At each step the agent receives only an observation \(o_{t} = \Omega(s_t) \in O\) and must select actions without direct access to the underlying state \(s_t\). In environments with randomized but unobserved parameters, the true state includes both the robot configuration and the latent environment parameters, making the problem partially observable even when the robot's own configuration is fully sensed.
We formulate loco-manipulation as a Markov Decision Process (MDP) \cite{sutton_reinforcement_2018}, defined by $(S, A, f, r, \gamma)$, where $S$ and $A$ denote the state and action spaces, $f$ the transition function, $r$ the reward function, and $\gamma$ the discount factor. The goal of reinforcement learning is to find parameters $\theta$ for $\pi_\theta(a|s)$ that maximize the expected discounted return
\begin{equation}
J(\theta) = \mathbb{E}_{\tau \sim \pi_\theta}\left[\sum_{t=0}^{T-1}\gamma^t r(s_t,a_t)\right],
\end{equation}
where $\tau=(s_0,a_0,s_1,\dots)$ denotes a trajectory induced by the policy.

When the full state is not observable, the problem becomes a Partially Observable MDP (POMDP) \cite{kaelbling1998planning}, defined by $(S,A,O,f,\Omega,r,\gamma)$, where $O$ is the observation space and $\Omega:S\rightarrow O$ is the observation function. At each step the agent receives an observation $o_t=\Omega(s_t)$ instead of the true state. In practice, randomized but unobserved environment parameters augment the true state, making the problem partially observable even when the robot’s configuration is fully sensed.

\subsection{Robust Locomotion Policy for Motion Prior Generation}

% In this subsection, we learn a robust locomotion policy that serves as a reference for the loco-manipulation policy described in the following subsection. However, learning a periodic and stable locomotion policy is challenging even in the locomotion-only scenario, as it typically requires extensive terrain and dynamics randomization to produce walking behaviors that generalize across diverse surface conditions. Because the policy relies solely on proprioceptive feedback, without explicit terrain or contact sensing, the randomized terrain properties and dynamics parameters remain unobserved and therefore act as latent variables. Consequently, we formulate the learning problem as a partially observable Markov decision process (POMDP). At each control step $t$, a partial observation $o_t$ is provided to the policy instead of the full system state $s_t$. The reference policy is therefore defined as a mapping $\pi_{\text{ref}} : \mathcal{O} \rightarrow \mathcal{A}$.
Learning a stable and robust locomotion policy that can handle multiple speeds and gait patterns is challenging, particularly when the underlying physics parameters are unknown, and the policy relies solely on proprioceptive observations. To improve robustness, training typically incorporates extensive terrain and dynamics randomization. Because these variations are not directly observable, we formulate the problem as a POMDP, where the policy receives a partial observation $o_t$ instead of the full system state $s_t$ at each control step. \\

% The reference locomotion policy is therefore defined as $\pi_{\text{ref}}:\mathcal{O}\rightarrow\mathcal{A}$.

\textbf{Observation.}
The policy receives an observation vector $o_t \in \mathbb{R}^{77}$ defined as
\begin{equation}
o_t =
\left[
\mathbf{q}_t,
\dot{\mathbf{q}}_t,
\mathbf{g}_t,
\omega_t,
\bar{\mathbf{v}}_t,
\bar{\omega}_t,
\bar{\beta}_t,
\bar{\mathbf{p}}^{\text{ee}}_t,
\bar{\mathbf{q}}^{\text{ee}}_t
\right],
\end{equation}
where $\mathbf{q}_t$ and $\dot{\mathbf{q}}_t$ denote the joint angles and joint velocities, respectively. The IMU measurements consist of the projected gravity vector $\mathbf{g}_t \in \mathbb{R}^3$ expressed in the robot base frame and the base angular velocity $\omega_t \in \mathbb{R}^3$ measured by the gyroscope. The remaining components correspond to command inputs: $\bar{\mathbf{v}}_t$ is the desired base linear velocity, $\bar{\omega}_t,$ is the desired base angular velocity, $\bar{\beta}_t$ is the heading command, and $\bar{\mathbf{p}}^{\text{ee}}_t$ and $\bar{\mathbf{q}}^{\text{ee}}$ are the end-effector position and orientation command specified relative to the robot base frame. Given the observation $o_t$, the reference policy $\pi_{\text{ref}}$ outputs target joint angles for both the leg and arm joints, i.e., $a_t \sim \pi_{\text{ref}}(a \mid o_t)$. Specific observation and command configurations are summarized in Tables~\ref{tab:observations} and~\ref{tab:command_ranges}.

\textbf{Reward.}
The reward function for training $\pi_{\text{ref}}$ encourages stable locomotion while accurately tracking commanded base velocity, heading, and end-effector targets and penalizes jerky or inefficient motions. The reward function is computed as 
\begin{equation}
    \begin{aligned}
        r = r_{\text{tracking}} + r_{\text{regularization}} + r_{\text{gait shaping}} + \mathbb I_{\text{term}}
    \end{aligned}
\end{equation}
where 
\begin{equation}
    \begin{aligned}
        r_{\text{tracking}} & = e^{{\|\bar{\mathbf{v}} - \mathbf{v}^{xy}\|^2}
        } + e^{{(\bar{\omega} - \omega)^2 }} + e^{{(\bar{\beta} - \beta)^2 }} \\
        & + {e^{\,\|\bar{\mathbf{p}}^{\text{ee}}  
        - \mathbf{p}^{\text{ee}}\|_1} + e^{-d_{\text{rot}}(\bar{\mathbf{q}}^{\text{ee}},\, \mathbf{q}^{\text{ee}})}} \\
    \end{aligned}
\end{equation}
\begin{equation}
    \begin{aligned}
        r_{\text{regularization}}  =  
         -\Big( \|\dot{\mathbf{a}}\|^2 + \|\boldsymbol{\tau}\|^2 & + \|\ddot{\mathbf{q}}\|^2 + \|\dot{\mathbf{q}}\|^2 + W \\ 
        + \sum \big( \big[\min(\mathbf{q} - \mathbf{q}^{\text{low}},\; 0)\big]^2 
        & + \big[\max(\mathbf{q} - \mathbf{q}^{\text{high}},\; 0)\big]^2 
        \big) \Big) \\ 
    \end{aligned}
\end{equation}
\begin{equation}
    \begin{aligned}
        r_{\text{gait shaping}}  = -\|\mathbf{q}^{\text{hip}}\|^2 & + e^{{\|\mathbf{g}\|^2 }} + e^{{(h - 0.31})^2} \\
        - (\sum \big(|\mathbf{q}^{\text{t}} - 0.5| - 0.4\big)^2 & + \big(|\mathbf{q}^{\text{c}} + 1.86| - 0.3\big)^2 )  \\ 
        -\|\mathbf{f}^{\text{f}}\| & -\sum_{\text{legs}} (|\mathbf{q} - \mathbf{q}^{\text{d}}|)\mathbb{I}_{\bar{\mathbf{v}}} \\
    \end{aligned}
\end{equation}
\noindent
where $d_{rot}(\cdot,\cdot)$ denotes the orientation distance between two quaternions, $\mathbf{q}^{\text{c}}$ are the calf joints, $\mathbf{q}^{\text{t}}$ are the thigh joints,  $\mathbf{q}^{\text{d}}$ is the default joint angle, $\mathbf{q}^{\text{low}}$ and $\mathbf{q}^{\text{high}}$ are the minimum and maximum joint angles, $W$ is the mechanical power, $h$ is the base height, $\mathbf{f}^\text{f}$ is the foot contact force, $\mathbb I_{\text{term}}$ indicates early termination, and $\mathbb{I}_{\bar{\mathbf{v}}}$ indicates whether the commanded velocity is greater than 0.1.

\begin{table}[t]
\vspace{0.5em}
\centering
\caption{Observation Specifications.}
\label{tab:observations}
\begin{tabular}{lc}
\toprule
\textbf{Observation} & \textbf{Noise Range} \\
\midrule
Base angular velocity & $[-0.20, 0.20]$ \\
Projected gravity & $[-0.05, 0.05]$ \\
Base velocity command & $[-0.10, 0.10]$ \\
Heading command & $[-0.10, 0.10]$ \\
End-effector pose command & $[-0.05, 0.05]$ \\
Relative joint positions & $[-0.02, 0.02]$ \\
Relative joint velocities & $[-1.50, 1.50]$ \\
Previous action & None \\
\bottomrule
\end{tabular}
\end{table}

\begin{table}[t]
\centering
\caption{Command ranges for reference locomotion policy.}
\label{tab:command_ranges}
\begin{tabular}{lc}
\toprule
\textbf{Command} & \textbf{Range} \\
\midrule
$\bar{\mathbf{v}}_x$ (Global Velocity) & $[0.0,0.0]\cup[0.1,1.0]$ \\
$\bar{\mathbf{v}}_y$ (Global Velocity) & $[0.0]$ \\
$\bar{\mathbf{\omega}}$ (Angular velocity) & Computed as error correction \\
$\bar{\beta}$ (Heading Target) & Computed as Error correction \\
$\bar{\mathbf{p}}^{\text{ee}}_x$ (EE position) & $[0.1,0.6]$ \\
$\bar{\mathbf{p}}^{\text{ee}}_y$ (EE position) & $[-0.4,0.4]$ \\
$\bar{\mathbf{p}}^{\text{ee}}_z$ (EE position) & $[0.15,0.4]$ \\
$\bar{\mathbf{q}}^{\text{ee}}_{\text{roll}}$ (EE orientation) & $[\pi/2]$ \\
$\bar{\mathbf{q}}^{\text{ee}}_{\text{pitch}}$ (EE orientation) & $[-0.1,0.1]$ \\
$\bar{\mathbf{q}}^{\text{ee}}_{\textbf{yaw}}$ (EE orientation) & $[0.0]$ \\
\bottomrule
\end{tabular}
\vspace{-1em}
\end{table}
\textbf{State Data Construction.}
After convergence, we roll out \(\pi_{\text{ref}}\) to collect a dataset of  state transition pairs
\begin{equation}
\mathcal{D}_{\text{ref}} =
\{(s^{\text{ref}}_t, s^{\text{ref}}_{t+1})\}_{t=1}^{T-1},
\end{equation}
where \(s^{\text{ref}}_t\) denotes the state at time step \(t\) generated by the reference policy.
This dataset captures the locomotion dynamics induced by \(\pi_{\text{ref}}\) and serves as reference motion data for partial adversarial imitation training.

\subsection{Loco-Manipulation Policy with Partial Adversarial Motion Priors}

In the second stage, we train a loco-manipulation policy $\pi$ assuming reliable cart position estimation. We leverage (i) a task reward designed for the cart-pushing objective and (ii) a style reward derived from a learned partial adversarial motion prior. To obtain a compact style reward, we adopt the Adversarial Motion Prior (AMP) framework~\cite{Peng_2021}, which trains a classifier to distinguish between reference motion data and synthesized motion. The locomotion policy obtained in the first stage is used as the reference motion. Note that we imitate reference state trajectories rather than action trajectories, since our goal is to preserve the locomotion style while allowing the actions to adapt to the additional cart-pushing task.

In standard AMP, the discriminator is trained using a full motion representation of both the reference motion and the motion generated by the learned policy. In our setting, however, forcing the policy to match the full reference state would unnecessarily constrain manipulation, since arm and end-effector states that are optimal for locomotion are not necessarily compatible with reaching and maintaining contact with the cart handle.

We therefore introduce \textit{partial AMP}, in which only a subset of the reference agent's state is imitated.
Concretely, we define a projection function $\phi(\cdot)$ that maps the full state $s_t$ to a lower-dimensional representation capturing only locomotion-relevant information. In our setting, $\phi(s_t)$ encodes the base motion together with lower-body joint kinematics, including the base angular velocity and the relative joint positions and velocities of the leg joints (hips, thighs, and calves). Manipulation-specific components, such as arm joint states and end-effector states, are excluded so that the adversarial objective regularizes locomotion style while allowing the upper body to adapt to the cart-pushing task. The discriminator is trained on projected state tuples $(\phi(s_t), \phi(s_{t+1}))$, ensuring that the imitation signal constrains only the base and lower-body motion.

We train a discriminator \(D_{\psi}(s)\) using a least-squares adversarial objective \cite{mao2017squaresgenerativeadversarialnetworks}, where reference transitions are labeled \(+1\) and policy-generated transitions are labeled \(-1\). The training objective is as follows:
\begin{equation}
\begin{aligned}
\arg\min_{\psi} \mathbb{E}_{(s, s') \sim \mathcal{D}_{ref}} [(D_{\psi}(\phi(s), \phi(s')) - 1)^2]             \\
+ \mathbb{E}_{(s, s') \sim \pi_\theta(s,a)} [(D_\psi(\phi(s), \phi(s')) + 1)^2] \\
+ \frac{w_{gp}}{2}\mathbb{E}_{(s, s') \sim \mathcal{D}_{ref}} [\| \nabla_{\psi} D_{\psi}(\phi(s), \phi(s')) \|^2], 
\end{aligned}
\end{equation}
where the first term encourages the discriminator to output \(+1\) for reference transitions, the second term encourages it to output \(-1\) for policy-generated transitions, and the third term penalizes large parameter gradients of the discriminator on reference data. This parameter-gradient regularization acts as an additional stabilizer early in training by discouraging excessively sharp discriminator updates and reducing overfitting to the reference dataset \cite{meschederlars}. The discriminator output is converted into a style reward that encourages the target policy's projected transitions to match the reference distribution. 

The final loco-manipulation policy is trained using the weighted sum of \(r_{\text{style}}\) and \(r_{\text{task}}\). Each term is defined as follows:
\begin{equation}
\begin{aligned}
r_{\text{style}}(s_t, s_{t+1}) &= \\
\max &[0, 
1 - 0.25(D_\psi((\phi(s), \phi(s_{t+1})) - 1)^2],
\end{aligned}
\end{equation}
\begin{equation}
    \begin{aligned}
        r_{\text{task}} & = r_{\text{tracking}} + r_{\text{regularization}} \\
        & -\|\mathbf{q}^{\text{hip}}\|^2 - \|\mathbf{f}^{\text{f}}\| - (v^z)^2 + e^{-40q^{\text{grip}}}\\
        & -\sum_{\text{legs}} (|\mathbf{q} - \mathbf{q}^{\text{d}}|)\mathbb{I}_{\mathbf{v}}  - \sum_{\text{feet}}\|\mathbf v\|_2  + \mathbb I_{\text{cart}} + \mathbb I_{\text{term}}
    \end{aligned}
\end{equation}

% \begin{equation}
% \begin{aligned}
% % \begin{array}{l}
% r_{\text{task}} =
% &\underbrace{
% e^{-16{\|\bar{\mathbf v}-\mathbf v^{xy}\|^2}}
% +
% e^{-4{(\bar{\omega}-\omega^z)^2}}
% +
% e^{-16{(\bar{\beta}-\beta)^2}}
% }_{\text{root tracking}}
% \\
% +
% &\underbrace{
% e^{-3\|\bar{\mathbf p}^{\text{ee}}-\mathbf p^{\text{ee}}\|_1}
% +
% e^{-d_{\text{rot}}(\bar{\mathbf q}^{\text{ee}},\mathbf q^{\text{ee}})}
% +
% e^{-40q^{\text{grip}}}
% }_{\text{arm tracking}}
% \\
% -
% &\underbrace{(
% \|\dot{\mathbf a}\|^2
% +
% \|\boldsymbol\tau\|^2
% +
% \|\ddot{\mathbf q}\|^2
% +
% \|\dot{\mathbf q}\|^2
% )}_{\text{motion regularization}}
% \\
% -
% &\underbrace{(
% (v^z)^2
% +
% \|\mathbf q^{\text{hip}}\|^2
% +
% \|\dot{\mathbf q}^{\text{arm}}\|^2
% +
% \mathbb I_{\text{term}}
% +
% \mathbb I_{\text{cart}}
% )}_{\text{penalties}}
% \\
% -
% &\underbrace{
% \sum
% \left(
% [\min(q-q^{\text{low}},0)]^2
% +
% [\max(q-q^{\text{high}},0)]^2
% \right)
% }_{\text{joint constraints}}
% \\
% +
% &\underbrace{
% e^{-25{\|\mathbf g\|^2}}
% -
% \sum_{\text{legs}}|q-q^{\text{def}}|
% -
% \sum_{\text{feet}}\|\mathbf v\|_2
% }_{\text{gait shaping}}
% \\
% % \end{array}
% \end{aligned}
% \end{equation}
\noindent
% \morgan{This is basically all the same as the prior reward list. Maybe just use r(loco) + manipulation specific rewards.}
where $\mathbb I_{\text{cart}}$ indicates cart height violation. The weights for the style reward and task reward are $1.75$ and $1.0$, respectively.
% \morgan{I removed the rest. It is already described in the prior section or in the observations.}
% \morgan{I renamed heading so that it wouldn't have the same name as the discriminator. I went through and tried to make all the notation consistent after Donghoon changed it, but you should make sure that it's right.}

% \begin{equation}
% \begin{aligned}
% r_{\text{task}} =\;&
% 65.0\, r_{\text{track-lin-vel}}
% + 7.5\, r_{\text{track-ang-vel}}
% + 6.0\, r_{\text{heading}}  \\
% &+ 20.0\, r_{\text{ee-pos}}
% + 10.0\, r_{\text{ee-ori}}  \\
% &- 1.0\, r_{\text{term}}
% - 0.35\, r_{\text{action-rate}}
% - 5.0{\times}10^{-6}\, r_{\tau}  \\
% &- 1.0{\times}10^{-5}\, r_{\text{dof-acc}}
% - 1.0\, r_{\text{lin-vel-}z}
% - 2.0\, r_{\text{hip}}  \\
% &- 0.025\, r_{\text{joint-limit}}
% - 5.0{\times}10^{-5}\, r_{\text{joint-vel}}  \\
% &+ 5.0\, r_{\text{flat}}
% - 0.4\, r_{\text{stand}}
% - 0.25\, r_{\text{feet-slide}} \\
% &- 0.25\, r_{\text{arm-joint-vel}}
% - 5.0\, r_{\text{cart-height}} .
% \end{aligned}
% \label{eq:full_reward}
% \end{equation}
% \(w_{\text{style}}\) controls the strength of the motion prior.
% \sehoon{Same here- better to define them using terms like q, qdot, tau}

We use a curriculum learning strategy for cart loco-manipulation. Training starts with straight-line cart pushing using short episodes for stable contact learning, followed by curved-path cart-turning with longer episodes to handle more complex coordination.

% and \(\alpha\) is the linear interpolation factor between task reward and style reward. Training alternates between collecting rollouts from \(\pi_\theta\), updating the discriminator using Eq. (5), and updating \(\pi_\theta\) with DRL under the combined reward in Eq. (7). 
% \todo{include architecture img}

% \begin{figure}
%     \centering
%     \includegraphics[width=1\linewidth]{ArchitectureDiagram.jpg}
%     \caption{Architecture \morgan{All figures should be pdfs so that they are vector graphics and do not blur when zoomed in.}}
%     \label{fig:architecture}
% \end{figure}

\begin{table}[tb]
\vspace{0.65em}
\centering
\caption{Domain randomization parameters for reference locomotion policy
training.}
\label{table:randomization}
\renewcommand{\arraystretch}{1.15}
\resizebox{\columnwidth}{!}{%
\begin{tabular}{@{}llll@{}}
\toprule
\textbf{Category} &
\textbf{Parameter} &
\textbf{Range} &
\textbf{Type} \\
\midrule
\multirow{3}{*}{Contact material}
  & Static friction          & $[0.4,\; 1.2]$                & Absolute \\
  & Dynamic friction         & $[0.2,\; 1.0]$                & Absolute \\
  & Restitution              & $[0.0,\; 0.2]$                & Absolute \\
\midrule
\multirow{2}{*}{External disturbance}
  & Base force (per axis)    & $[-10.0,\; 10.0]$\,N          & Absolute \\
  & Base torque (per axis)   & $[-2.0,\; 2.0]$\,Nm         & Absolute \\
\midrule
\multirow{3}{*}{Dynamics}
  & Link masses              & $[0.9,\; 1.1]$                & Scale \\
  & Joint stiffness (P gain) & $[0.9,\; 1.1]$                & Scale \\
  & Joint damping (D gain)   & $[0.9,\; 1.1]$                & Scale \\
\midrule
\multirow{4}{*}{Initial state}
  & Base position $(x,\,y)$  & $\pm\,0.05$\,m                & Offset \\
  & Base yaw                 & $\pm\,0.1$\,rad               & Offset \\
  & Base linear vel.\ $(x,\,y)$ & $\pm\,0.05$\,m/s          & Offset \\
  & Joint positions          & $\pm\,0.1$\,rad               & Offset \\
  & Joint velocities         & $\pm\,0.1$\,rad/s             & Offset \\
\bottomrule
\end{tabular}%
}
\end{table}

\begin{table}[tb]
\centering
\caption{Terrain randomization configuration.}
\label{tab:terrain}
\renewcommand{\arraystretch}{1.0}
\setlength{\tabcolsep}{4pt} % reduce horizontal padding
\begin{tabular}{lccc}
\toprule
\textbf{Type} & \textbf{Prop.} & \textbf{Param} & \textbf{Range} \\
\midrule
Grid boxes    & 0.42 & Height      & 0.05--0.10\,m \\
Rough terrain & 0.24 & Noise amp.  & 0.02--0.10\,m \\
Sine waves     & 0.17 & Wave amp.   & 0.02--0.08\,m \\
Pyramids      & 0.17 & Slope grade & 0.0--0.2 \\
\bottomrule
\end{tabular}
\vspace{-1.6em}
\end{table}

\section{Trajectory Following via Feedback Control}

To enable long-horizon path following for cart pushing, we employ a high-level feedback controller that generates path-following commands for the low-level loco-manipulation policy. Specifically, we use the Stanley controller~\cite{stanley}, a widely adopted method for autonomous driving, to compute a commanded forward velocity $\bar{\mathbf{v}}_x$ and yaw rate $\bar{\omega}$ at each control step.

The reference path is represented as a differentiable planar curve in the world frame. The target forward velocity is determined using a heuristic that reduces speed when tracking errors increase or when the path curvature $\kappa$ is large. At each control step, the controller finds the closest point on the path to the current cart position and computes the cross-track error $e_{\text{tr}}$ and heading error $e_{\text{head}}$. The target yaw rate is computed using the standard Stanley control law

\begin{equation}
\omega_{\text{target}}
=
- k_{\theta}e_{\text{head}}
-
\arctan\left(\frac{k_e e_{\text{tr}}}{v_{\text{target}}}\right),
\end{equation}
where $k_{\theta}$ and $k_e$ are controller gains. Finally, the target velocity and yaw rate are smoothed and clipped before being passed to the low-level policy as commands. Additional implementation details follow the standard Stanley controller formulation in~\cite{stanley}.

% \section{Trajectory Following via Feedback Control}

% The goal of generating a cart-pushing policy is to deploy to long-horizon path following scenarios such as navigating the aisles of a grocery store. As such, we utilize the Stanley Feedback Controller \cite{stanley}, commonly used in path following for autonomous vehicles, to generate high-level commands for the low-level loco-manipulation controller that solve the path-following problem.  At each control step, the controller outputs a forward speed (\(v_x\)) and a yaw rate (\(\omega\)), which are then passed as commands to the low-level policy.

% The reference path is a 2D curve in the world frame, such that the path is projected onto the XY-plane. It is represented in its parametric form
% $$
% \mathbf{p}(t)=
% \begin{bmatrix}
%     x(t)\\
%     y(t)
% \end{bmatrix} = 
% \begin{bmatrix}
% t\\
% f(t)
% \end{bmatrix}, \eqno{(8)}
% $$
% where 
% $$ \qquad t\in[t_{\min},t_{\max}], $$ 
 
% with analytical first and second derivatives
% $$
% \mathbf{p}'(t) = 
% \begin{bmatrix}
%     1 \\
%     f'(t)
% \end{bmatrix}, 
% \mathbf{p}''(t) = 
% \begin{bmatrix}
%     0 \\
%     f''(t)
% \end{bmatrix}.
% $$
% In our experiments, \(f(t)\) is a piecewise function. Each piece of the piecewise function must be differentiable. Thus, our assumption is that the path can always be approximated by some combination of differentiable functions. 

\begin{table*}[!ht]
\vspace{0.5em}
\centering
\renewcommand{\arraystretch}{1.15}
\caption{Baseline Comparisons with 95\% Confidence Interval.}
\label{tab:amp_ablations}
\resizebox{\textwidth}{!}{%
\begin{tabular}{llcccccc}
\toprule
\textbf{Env} &
\textbf{Method} &
\textbf{Survival Rate (\%)} &
\textbf{Lin Vel Error} &
\textbf{Ang Vel Error} &
\textbf{Heading Error} &
\textbf{EE Pos Error} &
\textbf{EE Ori Error} \\
\midrule
\multirow{4}{*}{\textbf{IsaacLab}} &
No AMP & 
$96.8 \pm 0.7$ &
$\mathbf{0.041 \pm 0.001}$ &
$\mathbf{0.041 \pm 0.001}$ &
$\mathbf{0.028 \pm 0.000}$ &
$\mathbf{0.020 \pm 0.000}$ &
$0.853 \pm 0.006$ \\

& Partial AMP (ours) &
$\mathbf{99.9 \pm 0.1}$ &
$0.045 \pm 0.000$ &
$0.085 \pm 0.001$ &
$0.035 \pm 0.001$ &
$0.038 \pm 0.000$ &
$\mathbf{0.293 \pm 0.001}$ \\

& Full AMP &
$97.2 \pm 0.7$ &
$0.070 \pm 0.003$ &
$0.072 \pm 0.003$ &
$0.047 \pm 0.003$ &
$0.027 \pm 0.001$ &
$0.827 \pm 0.005$ \\

& Hierarchical RL &
$97.5 \pm 0.7$ &
$0.088 \pm 0.005$ &
$0.092 \pm 0.002$ &
$0.110 \pm 0.001$ &
$0.039 \pm 0.001$ &
$0.338 \pm 0.002$ \\
\midrule
\multirow{4}{*}{\textbf{MuJoCo}} &
No AMP & 
$63.6 \pm 4.2$ &
$0.332 \pm 0.029$ &
$0.108 \pm 0.007$ &
$0.071 \pm 0.006$ &
$0.150 \pm 0.017$ &
$0.764 \pm 0.026$ \\

& Partial AMP (ours) &
$\mathbf{94.4 \pm 1.0}$ &
$\mathbf{0.156 \pm 0.009}$ &
$\mathbf{0.099 \pm 0.007}$ &
$\mathbf{0.052 \pm 0.003}$ &
$\mathbf{0.082 \pm 0.005}$ &
$0.465 \pm 0.018$ \\

& Full AMP &
$31.6 \pm 4.1$ &
$0.406 \pm 0.022$ &
$0.158 \pm 0.010$ &
$0.206 \pm 0.022$ &
$0.251 \pm 0.017$ &
$0.886 \pm 0.034$ \\

& Hierarchical RL &
$66.0 \pm 4.2$ &
$0.491 \pm 0.020$ &
$0.170 \pm 0.010$ &
$0.200 \pm 0.013$ &
$0.191 \pm 0.012$ &
$\mathbf{0.424 \pm 0.015}$ \\
\bottomrule
\end{tabular}%
}
\vspace{-1em}
\end{table*}
% \begin{table*}[tb]
% \centering
% \renewcommand{\arraystretch}{1.15}

\section{Experiments}
% \donghoon{here is what I suggest: Lists of questions that we want to answer through our experiments, A. Experimental Setup: e.g., simulation setup, robot, policy, etc B. Experimental Plan: B-1: Goal of the experiments,  B-2: Baselines B-3: Evaluation indexes. Results: highlight the most impressive results, data to support our claim and hypothesis - related to research questions, and discussion - e.g., deep analysis, limitation, etc}

\subsection{Experimental Design}
We study cart loco-manipulation in which the Unitree Go2 + WidowX AI pushes a shopping cart along a predefined path while maintaining continuous contact with the handle. The task is executed by a two-level controller: a high-level Stanley path-following controller that generates the commanded cart linear velocity and yaw rate, and a learned policy that outputs joint position targets to track these commands. The policy runs at 60\,Hz using joint position targets while the simulation runs at 240\,Hz.

The policies are trained using Proximal Policy Optimization (PPO)~\cite{schulman2017proximalpolicyoptimizationalgorithms} in massively parallelized IsaacLab simulation with \(2048\) environments. Both the actor and critic are multi-layer perceptrons with hidden sizes \([512,256,128]\). We use a learning rate of \(1\times10^{-4}\), discount factor $\gamma=0.99$, and clipping parameter $0.2$. Training converges in approximately two days on a single RTX 2070 SUPER. Domain and terrain randomization parameters are summarized in Tables~\ref{table:randomization} and~\ref{tab:terrain}, and latent environment parameters are re-sampled at every episode reset.

We train several variants under identical environment configurations and PPO hyperparameters:

\begin{itemize}
\item \textbf{No AMP:} A single-stage PPO loco-manipulation policy trained only with the task reward, without any imitation or style regularization.

\item \textbf{Partial AMP (ours):} A loco-manipulation policy trained with the proposed partial adversarial motion prior, where the AMP discriminator operates on a projection of the state to preserve lower-body motion style.

\item \textbf{Full AMP:} A loco-manipulation policy trained with a standard AMP objective applied to the full robot state, including both arm and end-effector states.

\item \textbf{Hierarchical Policy:} A modular baseline that uses the learned locomotion policy for lower-body control, while a separate policy learns residual arm actions for loco-manipulation. The lower-body actions are fixed by the locomotion policy and only the arm actions are learned for the cart-pushing task.
\end{itemize}

\begin{figure*}
\vspace{0.2em}
\centering
\includegraphics[width=15cm]{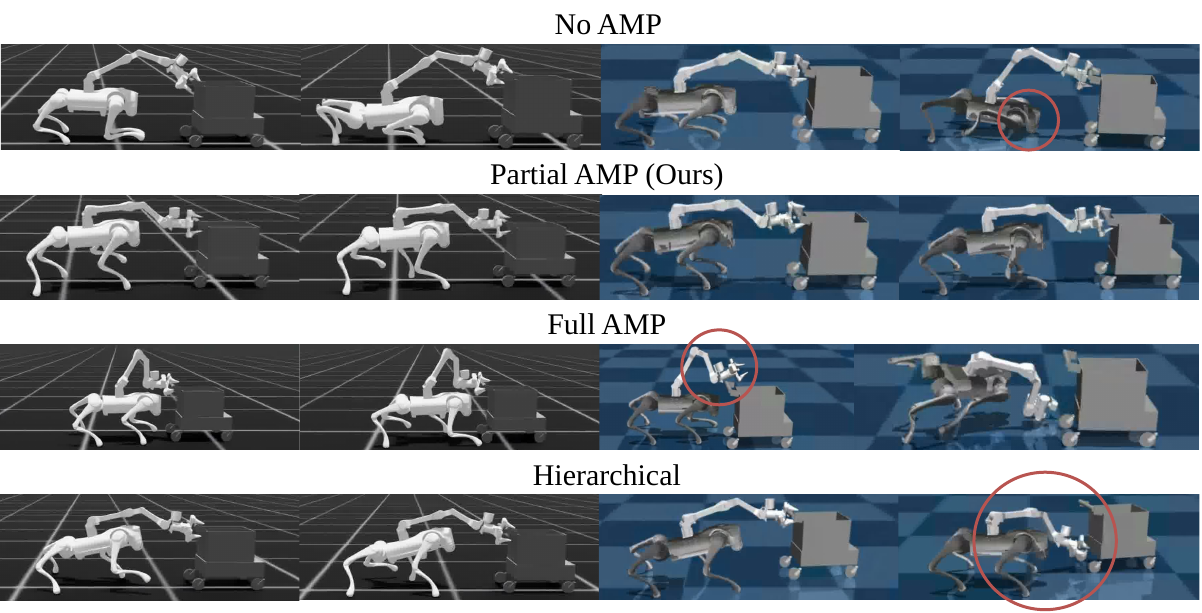}
\caption{\textbf{Sim-to-Sim Qualitative Comparison.} Depicted on the left is the policy deployed in the source simulator, IsaacSim, and on the right is the target simulator, MuJoCo. Our policy, partial AMP, demonstrates robustness when transferred, showcasing steady locomotion and end effector stability. On the other hand, the baselines are shown with their most common failure points. The No AMP baseline demonstrates feet slip, causing the head to collide with the ground. The Full AMP baseline demonstrates overfitting to the reference manipulator position instead of focusing on end effector tracking, which causes the end effector to slip off the handle. In hierarchical AMP, the robot adopts a conservative stance to prevent falling, resulting in high velocity tracking errors due to the lack of cart pushing.}
\label{fig:motion}
\vspace{-1em}
\end{figure*}

\vspace{-0.3em}
\subsection{Simulation Experiments}

% IsaacLab policies are tested on 2000 randomized environments. MuJoCo policies are tested on 500 randomized environments.
Table~\ref{tab:amp_ablations} and Fig.~\ref{fig:motion} compare the performance of the proposed and baseline methods in both IsaacLab and MuJoCo environments. The cart loco-manipulation task requires coordinating stable locomotion and accurate manipulation and is trained with a large number of reward terms. While all methods achieve high task success in the training environment (IsaacLab), they exhibit noticeably different qualitative motion patterns, leading to substantial differences in sim-to-sim transfer performance (MuJoCo). In summary, Partial AMP produces the most consistent and stable behaviors across environments.

In IsaacLab, all methods achieve high survival rates (approximately 97--100\%) and successfully perform the task. No AMP achieves low tracking errors but typically exhibits a bounding gait with less stable end-effector motion. In contrast, Partial AMP produces a more regular trotting gait with improved end-effector stability. Full AMP tends to overfit to the arm states observed during training, often keeping the arm close to the base,  resulting in unstable contact behaviors. The hierarchical policy achieves reasonable task performance but shows limited coordination between locomotion and manipulation due to the fixed lower-body policy.

After transfer to MuJoCo, performance differences become significantly larger. The most common failure mode across methods is loss of contact between the gripper and the cart handle, making stable end-effector motion critical for successful transfer. Overall, Partial AMP achieves the highest survival rate (94.4\%) and the lowest tracking errors by producing stable trotting motions and consistent end-effector behavior. No AMP achieves a substantially lower survival rate (63.6\%), reflecting the limited robustness of reward-only training due to its less stable bounding-like gait. Full AMP performs worst overall (31.6\%), as constraining the full-body motion leads to overly restricted arm behaviors and unstable end-effector motion (Fig.~\ref{fig:motion}, third row). The hierarchical policy achieves intermediate performance (66.0\%) but remains less stable, since the fixed locomotion policy cannot adapt when manipulation errors occur.

\vspace{-0.3em}
\subsection{Robustness}

To evaluate the robustness of the learned policies in a sim-to-sim transfer setting, we conducted stress tests across various physical parameters. Specifically, the controllers were evaluated against variations in four environmental properties: slide friction (ranging from 0.0 to 2.0), robot mass (scaled from 0.5 to 1.5), cart mass (scaled from 0.5 to 1.5), and wheel damping (ranging from 0.0 to 0.02). Each parameter was evaluated with 50 trials in MuJoCo, while all other parameters remained randomized. A particular parameter was considered successful if the policy achieved a mean survival rate of at least 30\% and maintained a mean linear velocity tracking error of 0.5 or less.

\begin{figure}[!tb]
\vspace{0.2em}
\centering
\includegraphics[width=\linewidth]{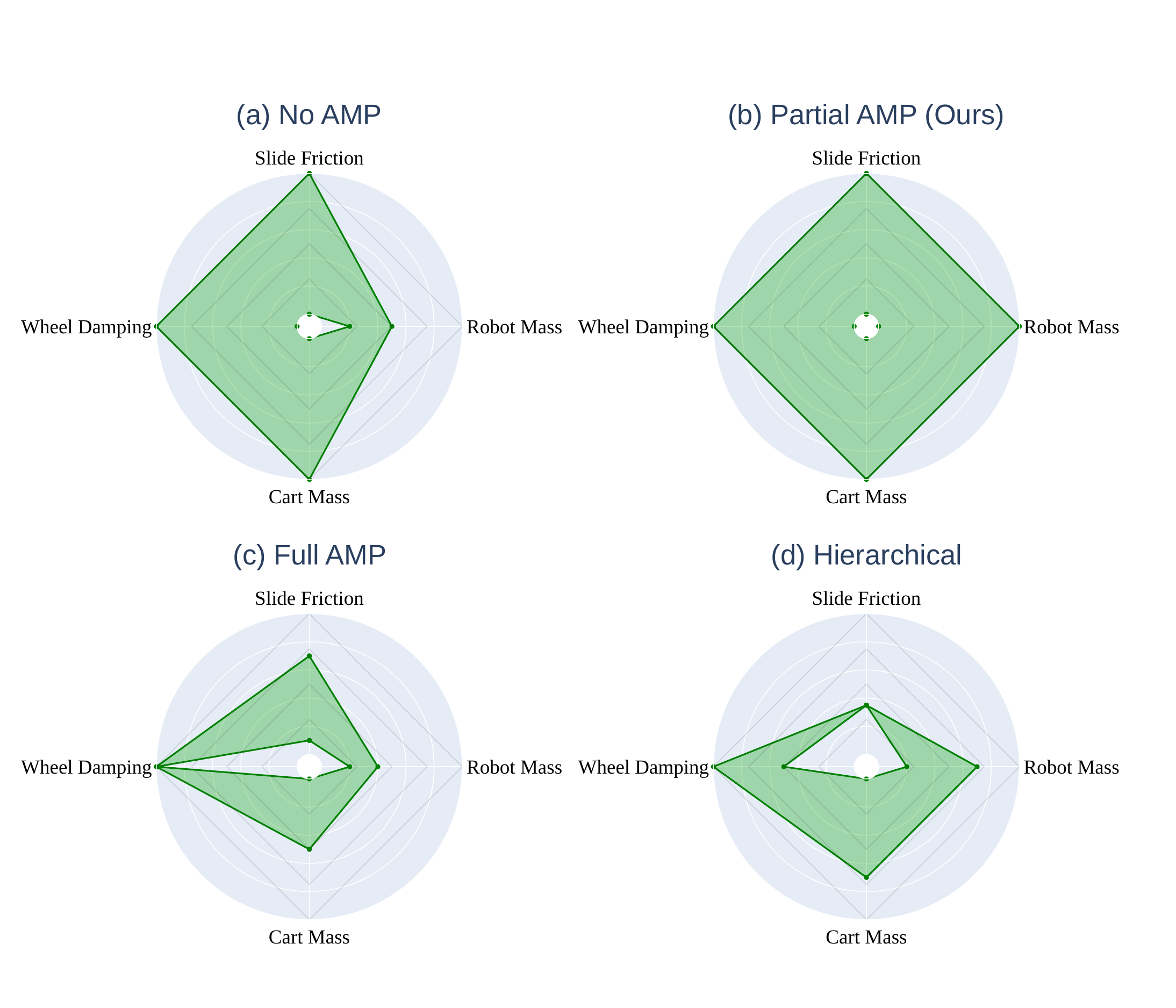}
\caption{\textbf{Robustness evaluation across environmental variations.} The shaded regions represent the parameter space where each policy satisfies the success criteria (survival rate $\ge 30\%$ and linear velocity error $\le 0.5$).}
\label{fig:robustness}
\vspace{-1em}
\end{figure}

The empirical results are illustrated in Fig.~\ref{fig:robustness}. The proposed Partial AMP framework exhibits the best robustness across all parameter variations.  No AMP demonstrates acceptable performance but is less resilient than Partial AMP under significant environmental shifts. Conversely, both the Full AMP and Hierarchical baselines fail to satisfy the success criteria under all tested conditions. The failure modes for these baselines are distinct: Full AMP predominantly suffers from premature episode termination and substantial tracking errors, whereas the Hierarchical policy avoids falling but yields excessively high velocity tracking errors, indicating an inability to actually push the cart. Furthermore, Partial AMP provides the most stable loco-manipulation behavior under dynamic variations. 

\vspace{-0.5em}

\subsection{Path Tracking}

% Measured in three different paths where the plot shows actual trajectory vs target trajectory and then another plot with the cross track error at that position. 
To evaluate long-horizon path-following capabilities, we tested the combined performance of the high-level feedback controller introduced in Section IV and the learned low-level Partial AMP policy. The integrated system was evaluated on three representative continuous trajectories: BeatPath, RiverPath, and SincPath. To rigorously assess the controller's ability to recover and adjust, environmental domain randomization was applied and the initial conditions of the robot were randomly sampled from a buffer of $\geq 8000$ states. 

\begin{figure}[!tb]
\vspace{0.2em}
\centering

\begin{subfigure}{0.7\linewidth}
    \centering
    \hspace{-2.5em}\includegraphics[width=\linewidth]{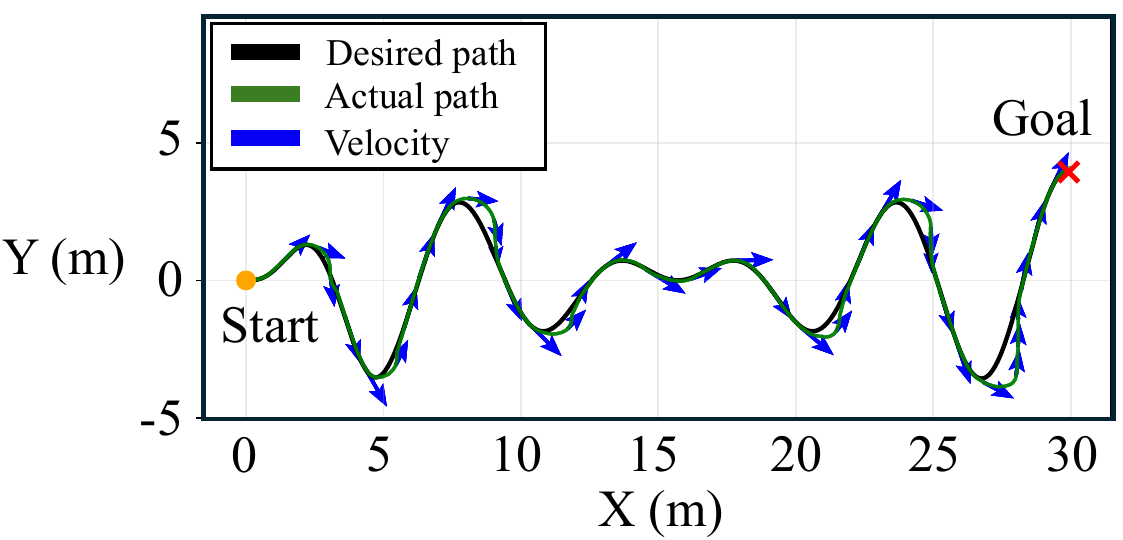}
    \caption{BeatPath}
\end{subfigure}
\begin{subfigure}{0.7\linewidth}
    \centering
    \hspace{-2.5em}\includegraphics[width=\linewidth]{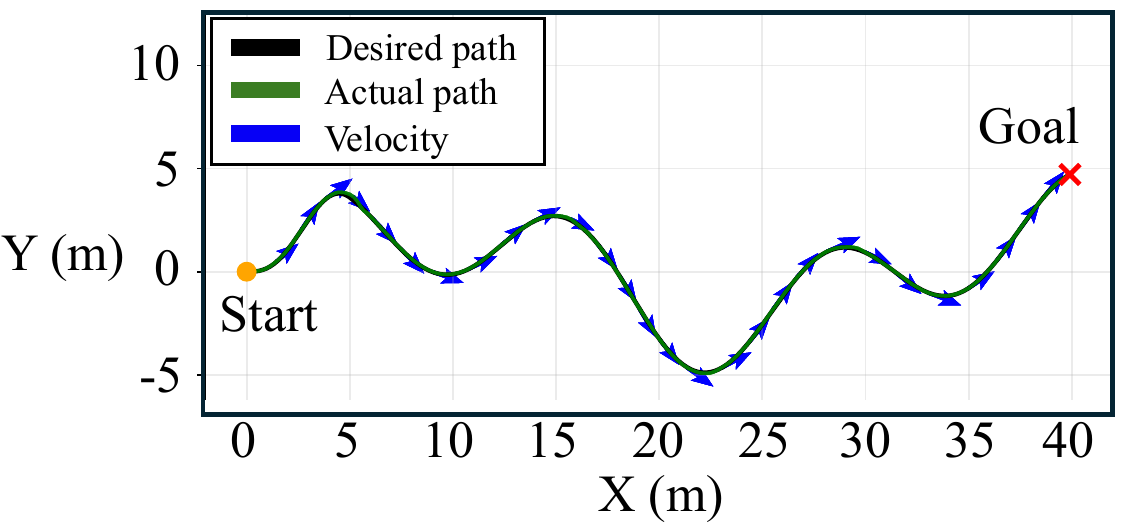}
    \caption{RiverPath}
\end{subfigure}
\begin{subfigure}{0.7\linewidth}
    \centering
    \hspace{-2.5em}\includegraphics[width=\linewidth]{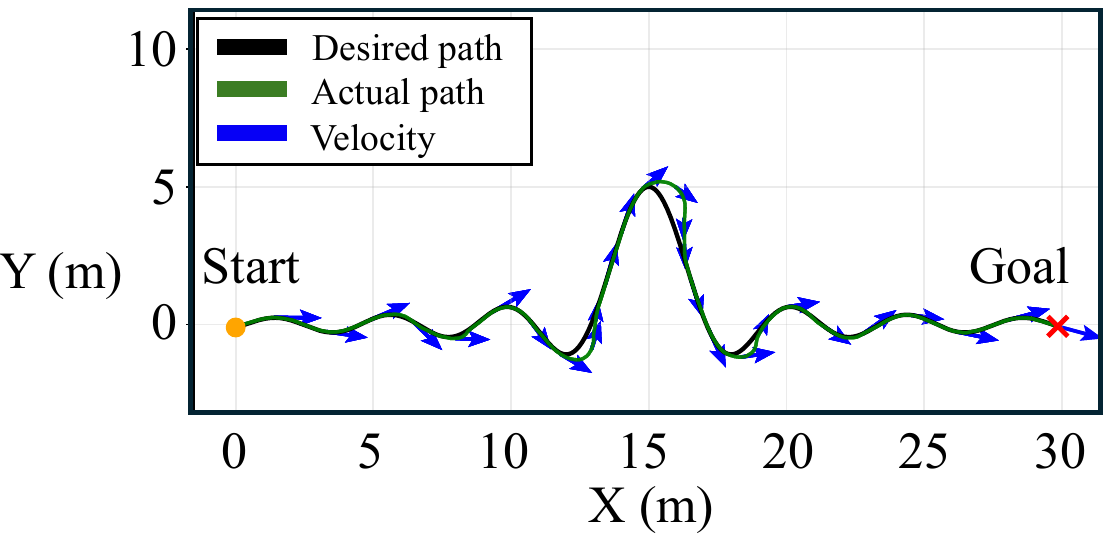}
    \caption{SincPath}
\end{subfigure}

\caption{
Individual runs of path-following performance on three representative trajectories. The mean cross-track error (averaged over five runs) is 0.1238 m for BeatPath, 0.0112 m for RiverPath, and 0.0885 m for SincPath, demonstrating accurate path tracking across different trajectory shapes. Blue arrows indicate the instantaneous velocity direction along the motion.}
\label{fig:path}
\vspace{-2em}
\end{figure}

The empirical path-following results are illustrated in Fig.~\ref{fig:path}. Across these varied spatial profiles, the combined framework demonstrated highly accurate tracking. Specifically, the mean cross-track errors (averaged over five runs) are 0.1238~m for BeatPath (Fig.~\ref{fig:path}a), 0.0112~m for RiverPath (Fig.~\ref{fig:path}b), and 0.0885~m for SincPath (Fig.~\ref{fig:path}c). We observe two primary sources of tracking error. First, initial deviations occur because the robot does not start exactly on the desired path due to the randomized initialization; however, the high-level controller successfully and rapidly compensates to converge onto the target trajectory. Second, further deviations are observed predominantly during sharp turns. These specific errors likely arise because navigating abrupt curves demands angular velocities that exceed the limits of the training distribution ($-0.8$ to $0.8$~rad/s), or they may reflect the inherent physical kinematic limitations of rapidly turning a shopping cart; additionally, the Stanley controller lacks lookahead capability, which limits its ability to anticipate sharp turns.

% \mili{Not sure if you want to include this, but the Stanley controller is also limited in that it has no lookahead ability, so the controller has no way of anticipating sharp turns. A different controller with a lookahead capability may also increase tracking accuracy.}

\section{Conclusion}
In this work, we presented a novel partial imitation learning framework for robust legged loco-manipulation applied to a cart-pushing task. By employing a partial adversarial motion prior, our approach preserves stable lower-body locomotion styles while allowing the upper body to adapt flexibly to manipulation objectives. Extensive experiments demonstrate that the proposed method outperforms baseline approaches in stability, path-tracking accuracy, and resilience to environmental variations, enabling reliable sim-to-sim transfer.

While our proposed Partial AMP framework demonstrates robust loco-manipulation along predefined trajectories, the current system does not incorporate visual feedback. Thus, a natural extension for future work is the integration of vision-based dynamic planning, which would enable legged manipulators to autonomously navigate unstructured environments, avoid obstacles, and generate trajectories in real time. Furthermore, as the primary failure mode observed across all evaluated methods is the unexpected loss of contact between the gripper and the handle, future research should address active contact recovery. Developing policies capable of detecting slip and autonomously executing re-grasping maneuvers will be crucial for deploying these systems in highly reliable, long-horizon real-world applications.

\vspace{-0.6em}

\bibliographystyle{IEEEtran}
\bibliography{root.bib}

\end{document}